\pdfoutput=1

\documentclass[11pt]{article}

\usepackage{emnlp2021}
\usepackage[linesnumbered,ruled]{algorithm2e}
\usepackage{times}
\usepackage{latexsym}
\usepackage{booktabs}
\usepackage{graphicx}
\usepackage{amsmath}
\usepackage{float}
\usepackage{xcolor}
\usepackage{adjustbox}
\usepackage{longtable}
\usepackage{subfig}

\usepackage{placeins} 
\usepackage{microtype} 
\usepackage{enumitem} 
\usepackage{amsmath, bm} 
\usepackage{comment} 

\usepackage[T1]{fontenc}

\usepackage[utf8]{inputenc}

\usepackage{microtype}
\title{BabyBear: Cheap inference triage for expensive language models}

\author{Leila Khalili, Yao You, John Bohannon \\
  Primer Technologies Inc. \\
  San Francisco, California \\
  \texttt{\{leila.khalili,yao.you,john\}@primer.ai}\\}

\begin{document}

\maketitle
\begin{abstract}
Transformer language models provide superior accuracy over previous models but they are computationally and environmentally expensive. Borrowing the concept of model cascading from computer vision, we introduce BabyBear, a framework for cascading models for natural language processing (NLP) tasks to minimize cost. The core strategy is inference triage, exiting early when the least expensive model in the cascade achieves a sufficiently high-confidence prediction. We test BabyBear on several open source data sets related to document classification and entity recognition. We find that for common NLP tasks a high proportion of the inference load can be accomplished with cheap, fast models that have learned by observing a deep learning model. This allows us to reduce the compute cost of large-scale classification jobs by more than 50\% while retaining overall accuracy. For named entity recognition, we save 33\% of the deep learning compute while maintaining an F1 score higher than 95\% on the CoNLL benchmark.
\end{abstract}

\section{Introduction}
"If all you have is a hammer, everything looks like a nail." This proverb describes a problem today for practitioners of Natural Language Processing (NLP). Our hammer is the transformer language model: BERT and its many descendants \cite{devlin2018bert,joshi2020spanbert, liu2019roberta}. Increasing the size, and thus cost, of these models continues in pursuit of ever better performance \cite{ghorbani2021scaling}. 

So why not solve everything with language models? Increasing computational complexity and memory consumption makes training these models harder and limits their scaling. Encoding text as high-dimensional vectors that pass through a neural network with hundreds of millions of parameters requires specialized hardware (GPUs, TPUs) that rapidly drains both your budget and the electrical grid \cite{xu2018scaling,vaswani2017attention,dozat2016deep}. Moreover, they simply are not always required. Less complex models like XGBoost or an LSTM are capable of solving portions of almost any given NLP task as well as a transformer based model can. For example, in Turkish sentiment analysis \cite{kavi2020turkish} XGBoost can achieve 88\% accuracy, suggesting that  88\% of the inference could be taken care of by XGBoost, leaving just 12\% for a deep learning model. The trick is the triage: How to identify those 88\% of data examples?

One strategy to reduce computation is to train a large model and then compress it. An example is DistilBERT, a smaller, faster version of BERT which can be fine-tuned. DistilBert has been found to be 60\% faster with 97\% of BERT's performance for some NLP tasks \cite{sanh2019distilbert}. However, DistilBERT is still a GPU-hungry deep learning transformer model.

Another strategy is to bypass unnecessary computation by early exiting, in which the inference is terminated when possible at an early layer of the neural network itself.  The early layers of BERT contain shallow features and perform poorly compared to a fine-tuned BERT with the same reduced number of layers \cite{li2020cascadebert}. Accordingly, CascadeBERT \cite{li2020cascadebert} has been proposed as a cascade of multiple BERT-like  models arranged in increasing depth of layers. TangoBERT \cite{mamou2022tangobert} is closely related to CascadeBERT but explores much smaller models such as distilled version of RoBERTa. Both CascadeBERT and TangoBERT still require expensive GPU computation due to their transformer architectures.

Our work uses a similar early exiting strategy, however  we use the cheapest options possible: Predictive models such as XGBoost that run on CPU at a fraction of the cost and latency. The idea of BabyBear in a nutshell: For any given task, the difficulty of data examples are not evenly distributed. To take advantage of this, we place a faster, cheaper, less powerful model (the "babybear") upstream of a slower, more expensive, more powerful model (the "mamabear"). The babybear model learns from mamabear as oracle. All that you, the human engineer, must do is choose a maximum accuracy cost you are willing to pay for the BabyBear computation savings. 

In this paper we introduce Babybear and apply it to several open source data sets and measure its impact on computational cost. Our code is available on github \footnote{\url{https://github.com/PrimerAI/primer-research/tree/main/babybear}} so you can start saving too.

\section{Methodology} \label{sec:method}

For the BabyBear algorithm we denote an ensemble of classification models $M_m$ where $m=\{b, p\}$.  The babybear model $M_b$ is a simple classification model or an ensemble of models such as logistic regression, random forest, and XGBoost \cite{brownlee2016xgboost}. The key requirement for the babybear model is the output of its prediction must include a confidence score. The mamabear model $M_m$ is the relatively more expensive and accurate model for which babybear is performing inference triage. We treat the predictions of mamabear as gold-labeled training data for babybear. 

Algorithm \ref{alg:inf} shows the procedure of inference triage where $x$ is the input document and $t$ is the confidence threshold.

\begin{algorithm}
    \SetKwInOut{Input}{Input}
    \SetKwInOut{Output}{Output}

    \underline{inference-triage}\;
    \Input{Models($M_m$, $M_b$), threshold, input($x$)}
    \Output{labels}
    \eIf{$M_{b}(x)$ > $t$}
      {
        return $M_b(x)$\;
      }
      {
        return $M_m(x)$\;
      }
    \caption{Inference triage}\label{alg:inf}
\end{algorithm}

\textbf{Confidence Threshold}. 
The first step  to find $t$ is to   choose a confidence function. This function maps probability distribution of each data point to a confidence measure. There are different methods to choose this confidence function. For example maximum probability in the predicted distribution \cite{huang2017multi,wang2020wisdom} or the gap between two top logits or entropy of the distributions \cite{streeter2018approximation}. In this paper we choose maximum probability for its simplicity.
Confidence threshold is a value in $[0, 1]$ that defines how confident $M_b$ must be to skip inference with $M_m$. The algorithm for choosing this parameter is described in Algorithm \ref{alg:conf}. X is the set of inputs $x_j$,  $A_f$ is the minimum performance the user demands from the final BabyBear system and $C$ is the confidence interval. 

\begin{algorithm}
    \SetKwInOut{Input}{Input}
    \SetKwInOut{Output}{Output}
    \underline{confidence-threshold}\;
    \Input{$X=\{x_j\}$, $A_f$, $C=\{c_j\}$}
    \Output{t}
    \For{$c$ in C:}{
    $A \leftarrow$ performance on $X$ at each $c_j$\;
    \If {$A \approx A_f$:}{
    Return $t \leftarrow c_j$
    }
    }
    \caption{Finding confidence threshold}\label{alg:conf}
\end{algorithm}

Higher threshold values mean that the babybear predictions are not trusted and most of the documents are sent to mamabear. This will reduce the savings to maintain performance. On the other hand, as the threshold approaches zero, most documents are handled by babybear which reduces GPU run time.

\subsection{Application examples}The general method described in  section \ref{sec:method} can be applied to different problems including classification and entity recognition tasks, relation extraction, etc. In this paper we focus on classification and entity recognition. Applying BabyBear on classification is very straightforward and can be followed as algorithm \ref{alg:inf}. In the case of NER, the algorithm needs some modifications which are described in the following section.

\begin{figure}[h]
    \centering
    \subfloat{\includegraphics[width=1\linewidth]{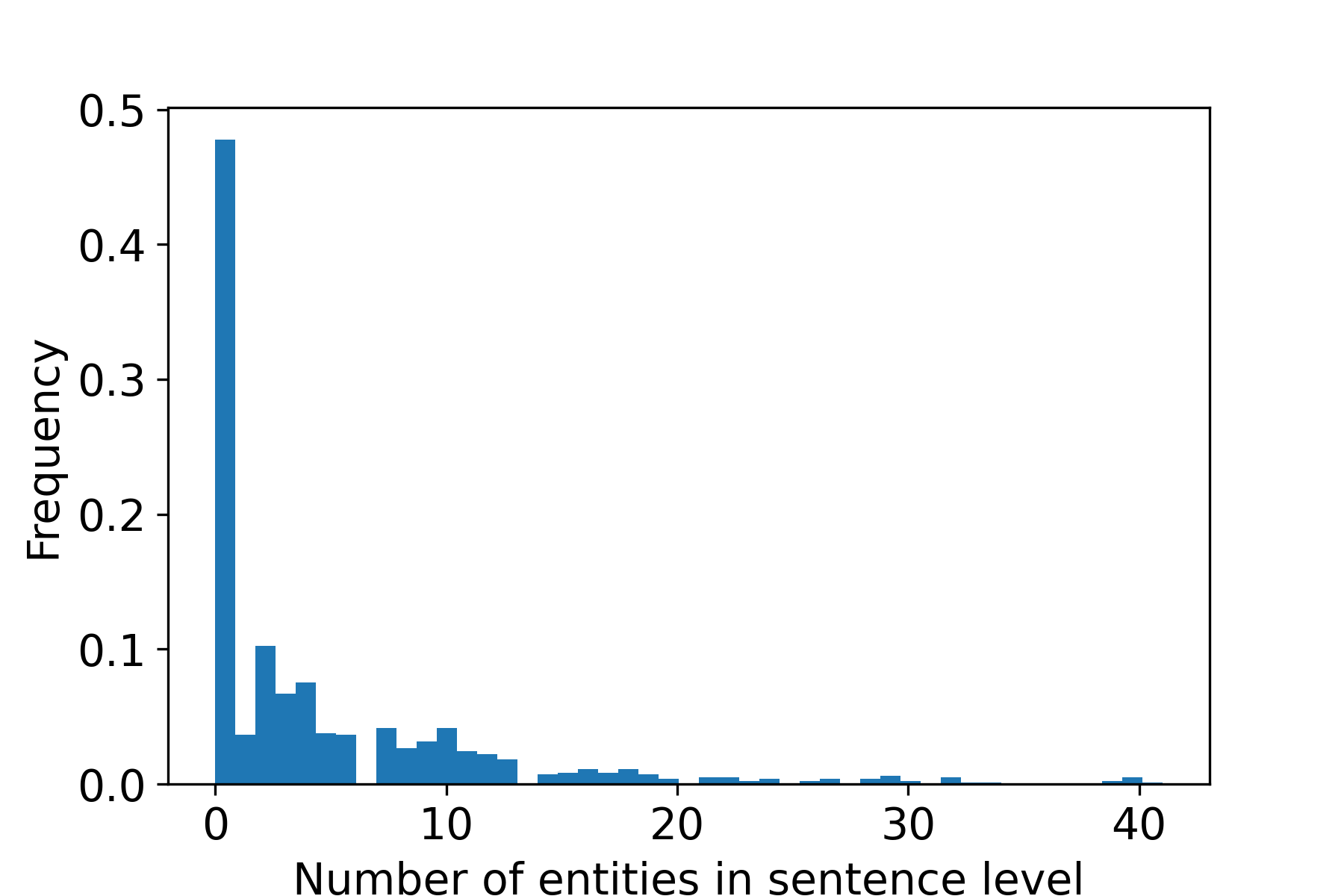}}\\
    \subfloat{\includegraphics[width=1\linewidth]{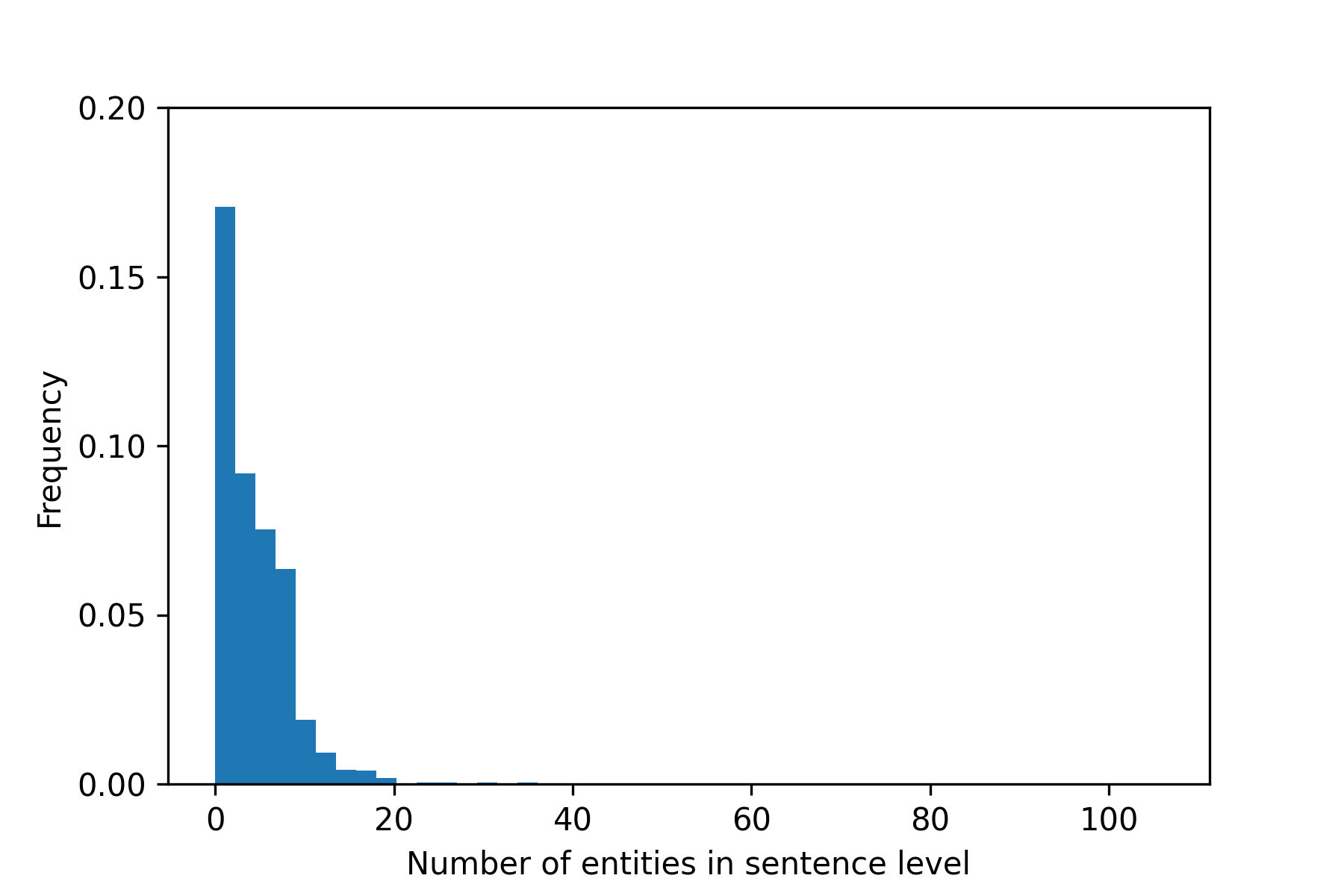}}
    \caption{Distribution of number of entities in each sentence in, top: ncbi\_disease with total number of 941 sentences, bottom: CoNLL data set with total number of 3454 sentences. }
    \label{fig:ncbi_dist}
\end{figure}

\textbf{BabyBear on entity recognition}. In typical English prose text, many sentences contain no relevant entities. At the sentence level for the ncbi\_disease data set \ref{fig:ncbi_dist}, 47\% of sentences have no biomedical entities. Therefore, if we can avoid sending these sentences to the mamabear model, we have the potential to save about half of the computational cost.

Accordingly, we cast the entity recognition babybear task as binary classification to identify two classes: sentences with "no-entities" and sentences "with entity". (For class "with-entity" there exists at least one entity). In the rest of our paper we refer to this method as \textbf{EntityBear}. In this scenario, high-confidence documents are identified as class "no-entity" and have $P_b(x)>t$. Unlike classification tasks in which the savings are associated with the high-confidence documents, in EntityBear the savings derive from the "no-entity" text we spare mamabear from seeing. Only "with-entity" text is sent to mamabear. Therefore, it is expected that the savings for EntityBear on entity recognition tasks will be lower than those for BabyBear on classification tasks.

We also expanded the EntityBear paradigm into  a cascade of multiple BabyBear models. These BabyBear models are EntityBear and DistilBear. We applied this to the CoNLL data set \cite{sang2003introduction} .  

The text input for both CoNLL and ncbi\_disease are single-sentence length, so documents must be split into sentences during pre-processing. The sentences classified "with-entities", are joined back together to form a document and passed to mamabear to extract entities. 
\section{Implementation and Experimental Setting}
\subsection{Tasks}\label{sec:data}
We conducted experiments on two text classification tasks (emotion recognition, sentiment analysis) and two entity recognition tasks (ncbi\_disease and CoNLL). See details below.

\textbf{Classification: Emotion Recognition}. This task consists of recognizing emotions expressed on Twitter. We used the data set from \cite{saravia2018carer} which is English Twitter messages with six basic emotions: anger, fear, joy, love, sadness, and surprise. These emotions are classified as classes from 0 to 5 respectively. This pipeline will download and cache a pretrained model bhadresh-savani/bert-base-uncased-emotion \footnote{\url{https://huggingface.co/bhadresh-savani/bert-base-uncased-emotion}} as the mamabear model.

\textbf{Classification: Sentiment Analysis}. The goal of sentiment analysis is to recognize if a tweet is positive, negative or neutral. We used the Semeval2017 data set from \cite{rosenthal2019semeval} which includes data from previous runs  (2013, 2014, 2015, and 2016) of the same SemEval task. The mamabear model is the pretrained model  cardiffnlp/twitter-roberta-base-sentiment-latest \footnote{\url{https://huggingface.co/cardiffnlp/twitter-roberta-base-sentiment-latest}} from huggingface.

\textbf{Entity recognition: ncbi\_disease}.
The ncbi\_data set is taken from \cite{dougan2014ncbi}. It contains the disease name and concept annotations of the NCBI disease corpus, a collection of 793 PubMed abstracts fully annotated at the mention and concept level to serve as a research resource for the biomedical natural language processing community. In this data set tag 0 indicates no disease mentioned, 1 signals the first token of a disease, and 2 the subsequent disease tokens. fidukm34/biobert\_v1.1\_pubmed-finetuned-ner-finetuned-ner \footnote{\url{https://huggingface.co/fidukm34/biobert_v1.1_pubmed-finetuned-ner-finetuned-ner}} is the mamabear model used for this data set.

\textbf{Entity recognition: CoNLL}.
The CoNLL-2003 data set  \cite{sang2003introduction} classifies four types of named entities: person, location, organization, and miscellaneous named entities that do not belong to the previous three groups. Here the mamabear model is dslim/bert-base-NER \footnote{\url{https://huggingface.co/dslim/bert-base-NER}} and the DistilBERT model is gunghio/distilbert-base-multilingual-cased-finetuned-conll2003-ner \footnote{\url{https://huggingface.co/gunghio/distilbert-base-multilingual\\-cased-finetuned-conll2003-ner?text=U.S+is+a+country}}.

\subsection{Experiment Results}

We applied inference triage to the data sets mentioned in section \ref{sec:data}.  For the classification tasks we used XGBoost \cite{brownlee2016xgboost} as babybear and Universal Sentence Encoder \cite{cer2018universal} as its embedding method. The code was run on a GPU instance of g4dn.2xlarge on Amazon Web Services (AWS).

The data set statistics are summarized in table \ref{tab:data set}. After training the babybear model, we applied algorithm \ref{alg:conf} on validation data set to find confidence threshold, $t$. The performance threshold is defined as  minimum accuracy of ${A_f=0.9}$ for classification task and ${A_f=0.99}$ for entity recognition tasks on the test data set. For the values of confidence threshold, saving and accuracy on test data set please refer to \ref{tab:data set}. Saving is the percentage of input data that are evaluated using the BabyBear model and are not sent to mamabear model. 

\begin{table*}[] 
\begin{tabular}{@{}llllllll@{}}
\toprule
Name & \#labels & Train(K) & Val.(K) & Test(K) & Threshold & \multicolumn{1}{r}{Saving} & \multicolumn{1}{c}{Accuracy} \\ \midrule
\begin{tabular}[c]{@{}l@{}}Emotion Recognition\end{tabular} & 6 & 11.2 & 4.8 & 2 & .825 & 64.6 & 86.9 \\
\begin{tabular}[c]{@{}l@{}}Sentiment Analysis\end{tabular} & 3 & 7 & 3 & 3 & .825 & 28.5 & 95.9 \\
ncbi\_disease & - & 5.4 & .9 & .9 & .785 & 9.6 & 99.2 \\
CoNLL & - & 7 & 3 & 3.4 & .978 & 18.3 & 97.8 \\ \bottomrule
\end{tabular}
\caption{Values of threshold, \% saving and accuracy for different data sets. }
\label{tab:data set}
\end{table*}

\subsubsection{Classification}
\begin{figure}[h]
    \centering
    \subfloat{\includegraphics[width=1\linewidth]{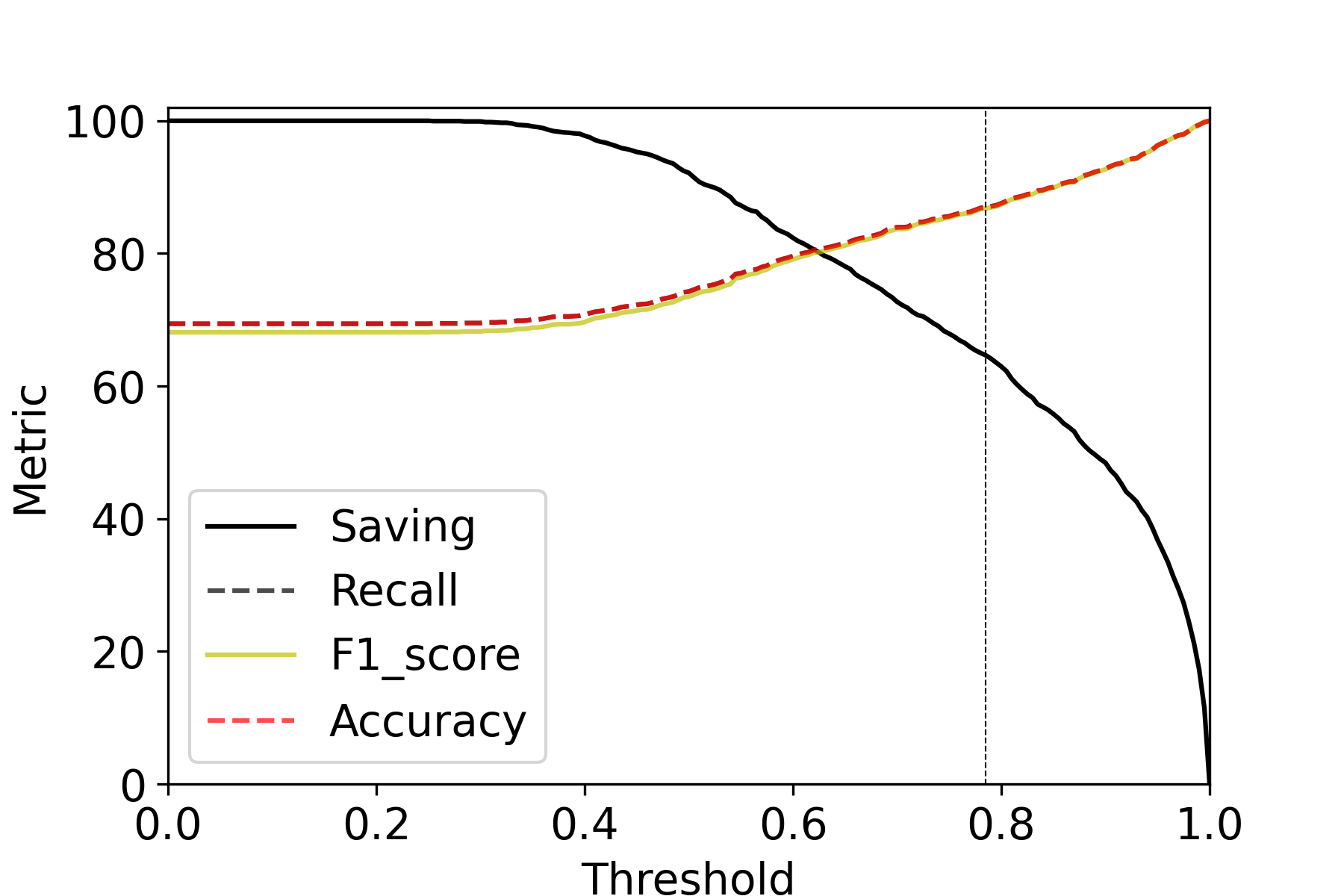}}\\
    \subfloat{\includegraphics[width=1\linewidth]{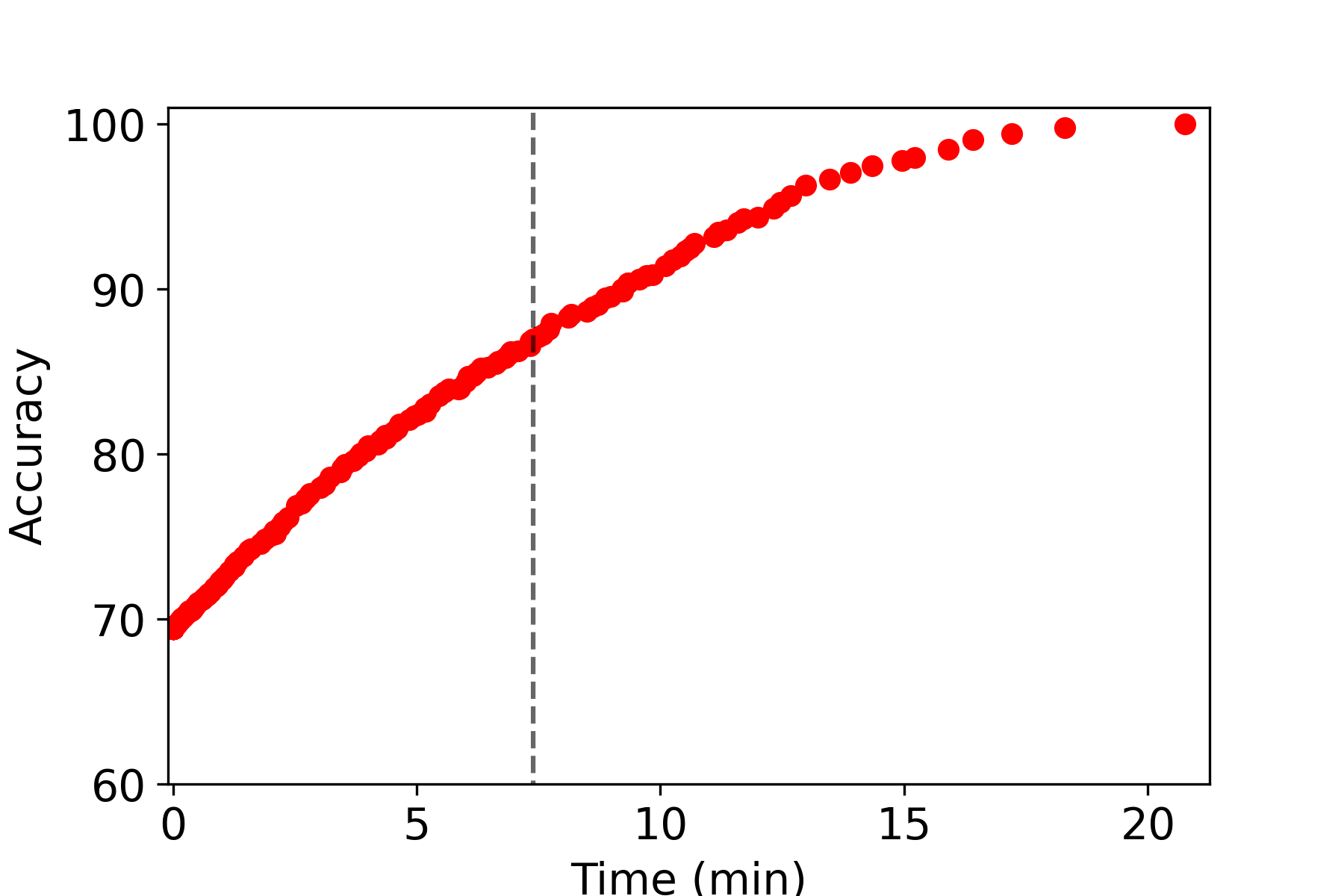}}
    \caption{Experiment on emotion data set. top: Accuracy and saving as a function of confidence threshold, bottom: Accuracy as a function of GPU run time for different values of confidence threshold.}
    \label{fig:emotion}
\end{figure}

\textbf{Emotion}. 
Figure \ref{fig:emotion}a shows the saving and accuracy as a function of threshold $t$. For the  chosen value of threshold $t$, over 64\% of documents are classified using the babybear model and the rest using the mamabear model. The overall accuracy of the inference triage model is 87\%. Figure \ref{fig:emotion}b shows the accuracy as a function of GPU run time. Each data point corresponds to a value of $t$. 

Running the entire data set on GPU with mamabear takes 20.76 mins, while applying BabyBear reduces the run time to 7.39 mins, a savings of 64\%.  \ref{fig:emotion}a).

\textbf{Sentiment Analysis}
Confidence threshold for this data set is $t=0.825$. About 28\% of the documents are processed by babybear with a system accuracy of 96\%. The savings with an accuracy closer to 90\% ($t=.725$) is about 50\%.

\begin{figure}[h]
    \centering
    \subfloat{\includegraphics[width=1\linewidth]{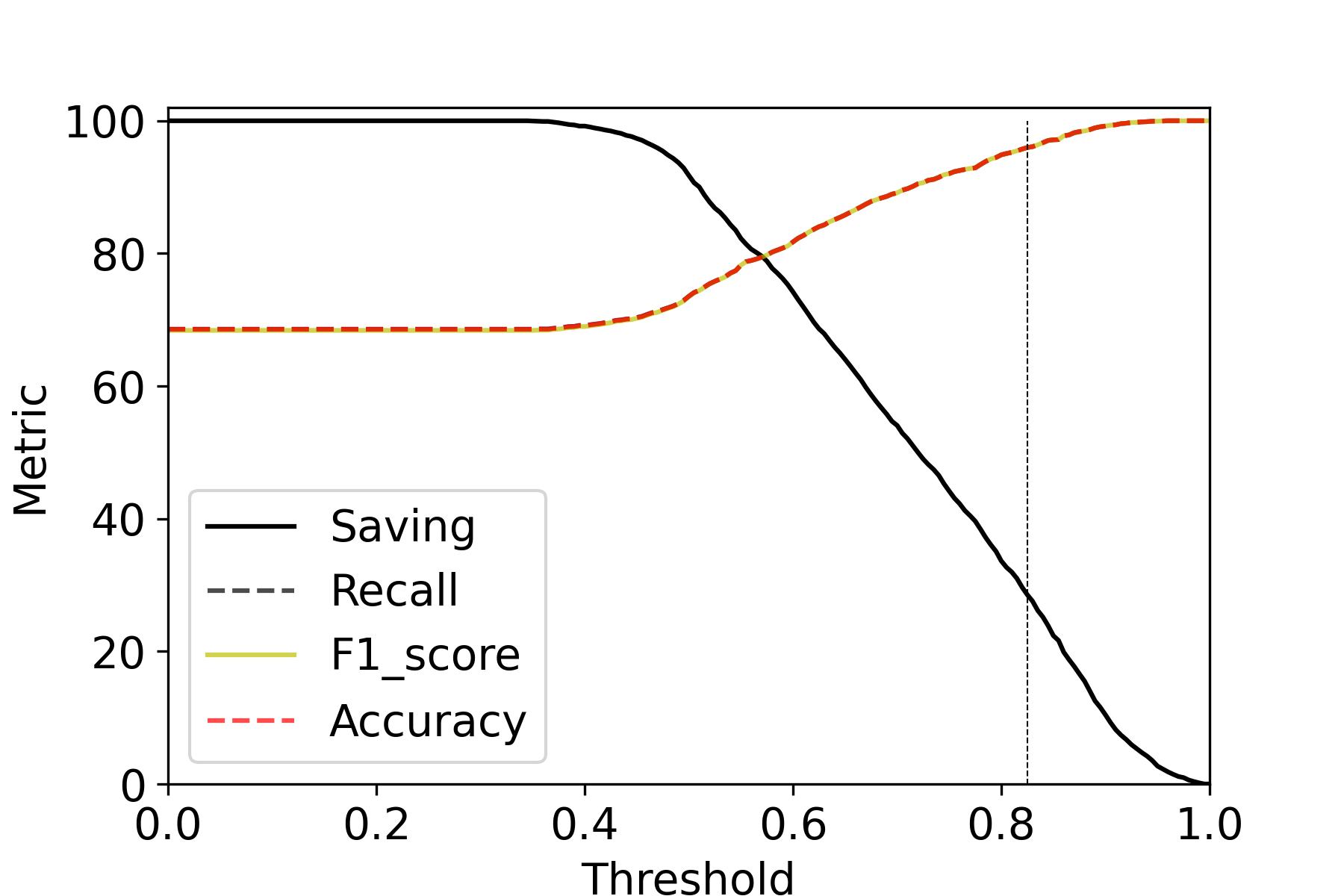}}\\
    \subfloat{\includegraphics[width=1\linewidth]{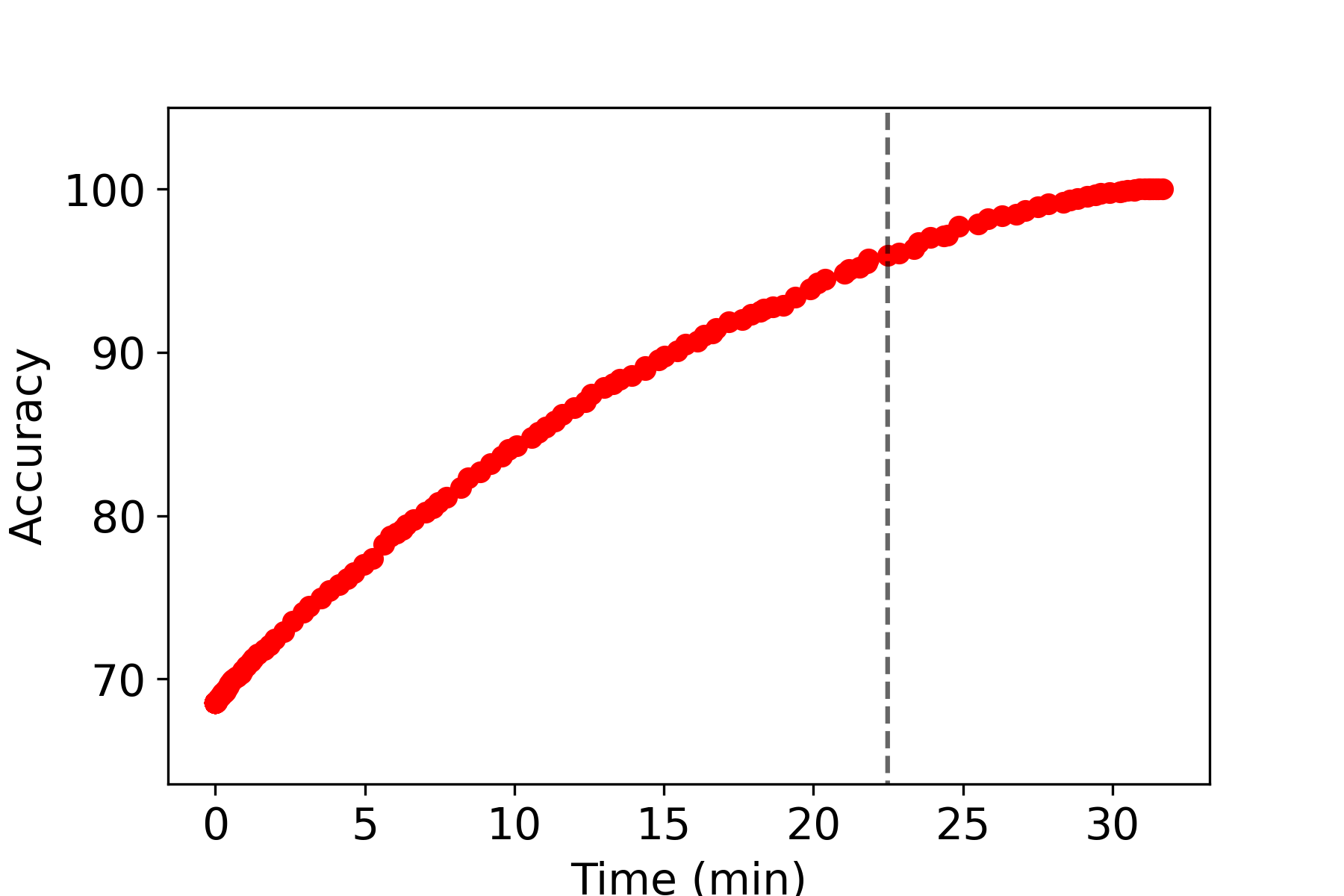}}
    \caption{Experiment on sentiment analysis. top: Accuracy and saving as a function of confidence threshold, bottom: Performance as a function of GPU run time for different values of confidence threshold.}
    \label{fig:sentiment}
\end{figure}

Maximum time is 31.35 min with mamabear alone which is reduced to 22.48 min after applying BabyBear, about 39\% savings of time. In this data set there is a difference between the savings in figure \ref{fig:sentiment} (top) and the time saving in \ref{fig:sentiment} (bottom). The reason is that the computational cost of each document is different. 

\subsubsection{Entity Recognition} \label{sec:ner}
Table \ref{tab:data set} shows that savings for the entity recognition tasks are less than those for classification tasks. The maximum possible savings in classification is 100\% of documents, while in entity recognition the maximum possible savings are defined by the ratio of "no-entity" versus "with-entity" in the text. Based on table \ref{tab:data set} the maximum savings is 47\% and 18\% for ncbi\_disease and CoNLL, respectively.

\begin{figure}[h]
    \centering
     \subfloat{\includegraphics[width=1\linewidth]{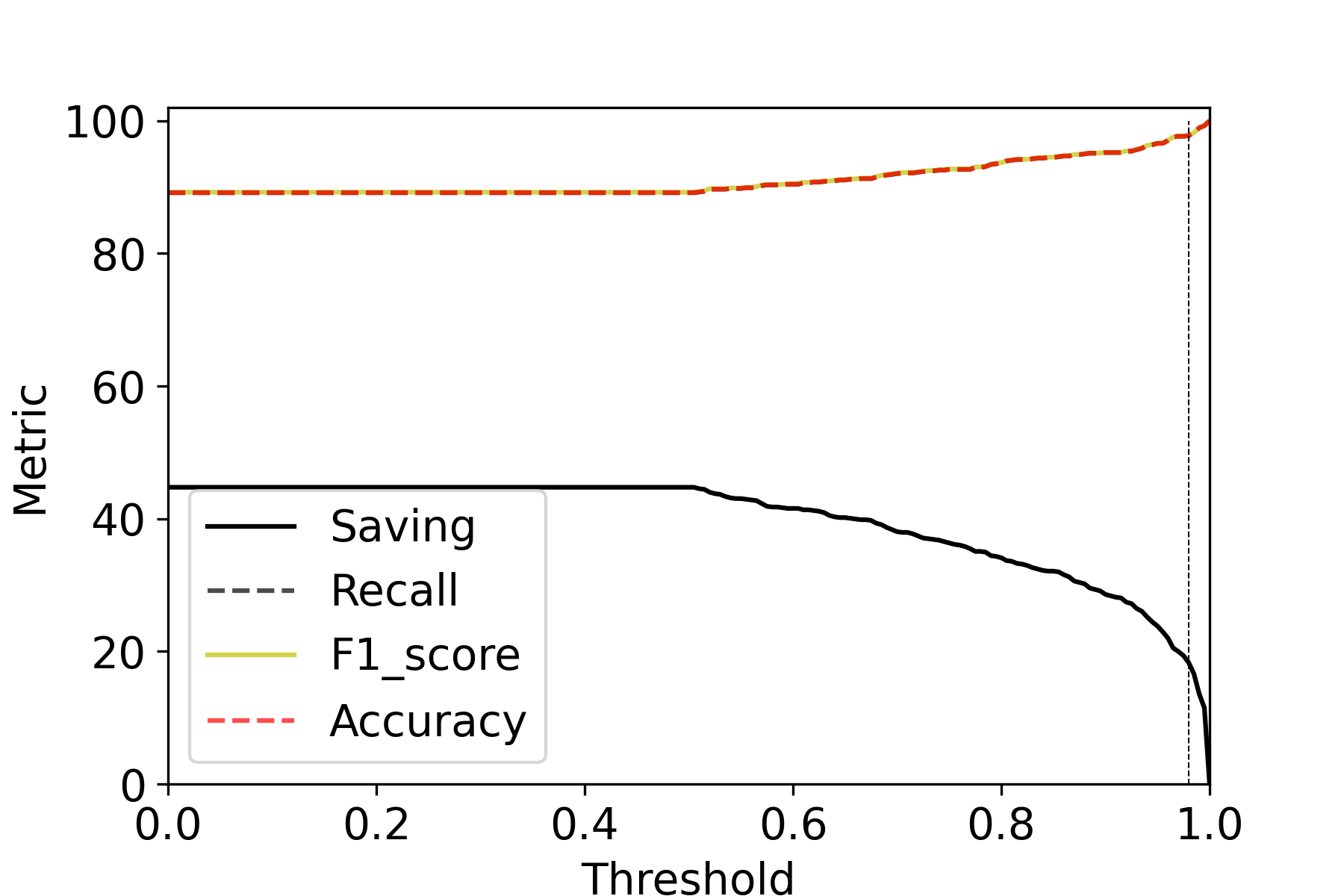}}
    \caption{ Performance and saving as a function of confidence threshold on ncbi\_disease data set.
    }
    \label{fig:ncbi_time}
\end{figure}

\textbf{ncbi\_disease}. Figure \ref{fig:ncbi_time} shows the performance and saving as a function of $t$. As it can be seen the  saving  is about 18\% with $t=.978$. 


In this data set GPU run time savings is about 17\% which is very close to the maximum possible savings  ( 18\%  saving shown in top figure \ref{fig:ncbi_time}). Since all of the documents confidently predicted to be class "with-entity" are sent to mamabear, all of the 18\% saving of GPU run time are from class "no-entity". The accuracy of the babybear model limits the achievable savings from inference triage for this task. About half of the documents could be classified as "no-entity" and spared from mamabear. Increasing the babybear model's accuracy is key to improving the savings.

\textbf{CoNLL}. In this data set  645 documents  out of 3453 have "no-entity" which is around 19\% of all the input data. This is the upper bound on savings. In this data set we save over 9.7\% of GPU run time which is very close to the percentage of documents not sent to mamabear (9.6\%  saving). Therefore, we are saving on 9\% of class "no-entity" which is close to half of the maximum possible saving. 

\begin{figure}[h]
    \centering
    \subfloat{\includegraphics[width=1\linewidth]{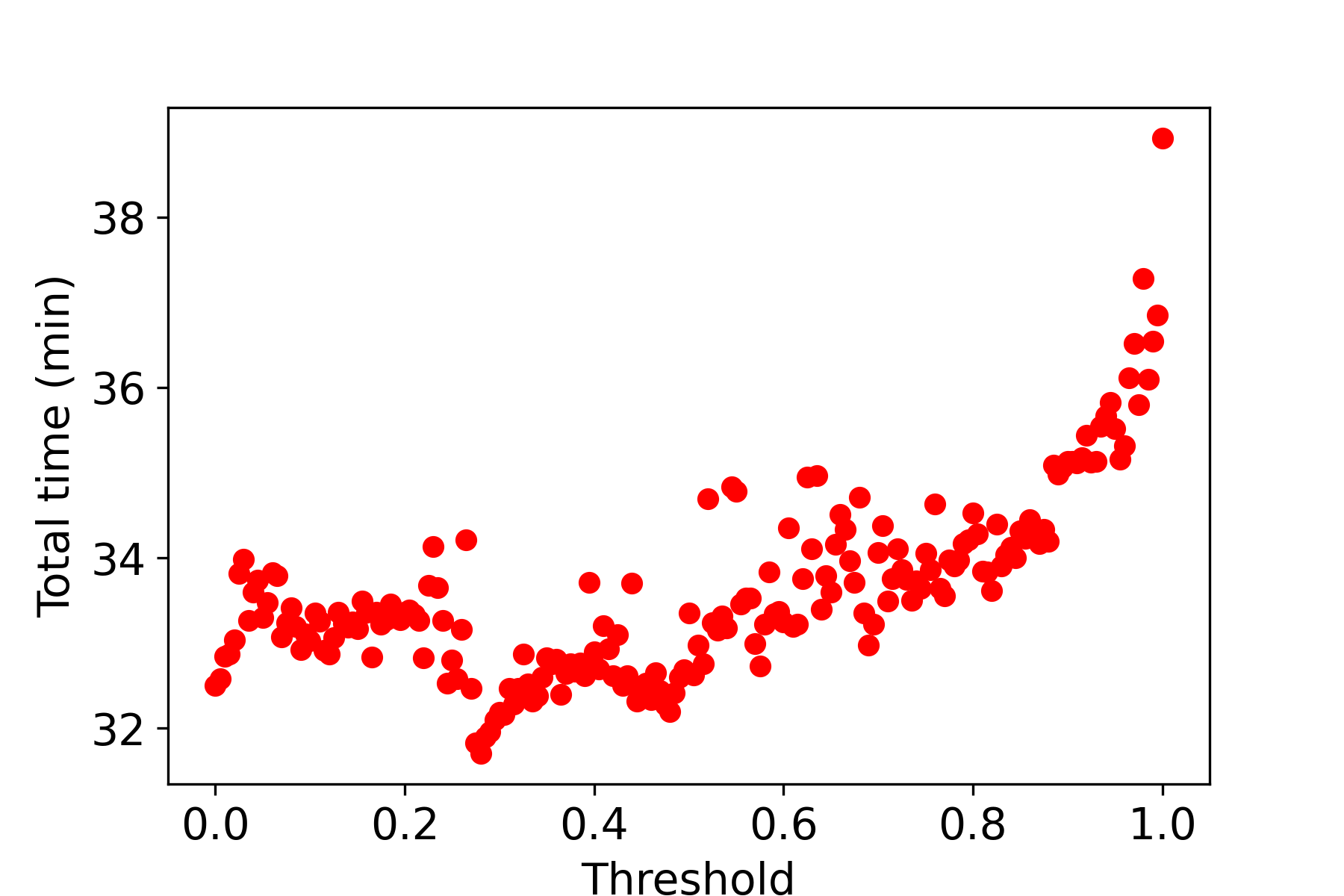}}\\
    \subfloat{\includegraphics[width=1\linewidth]{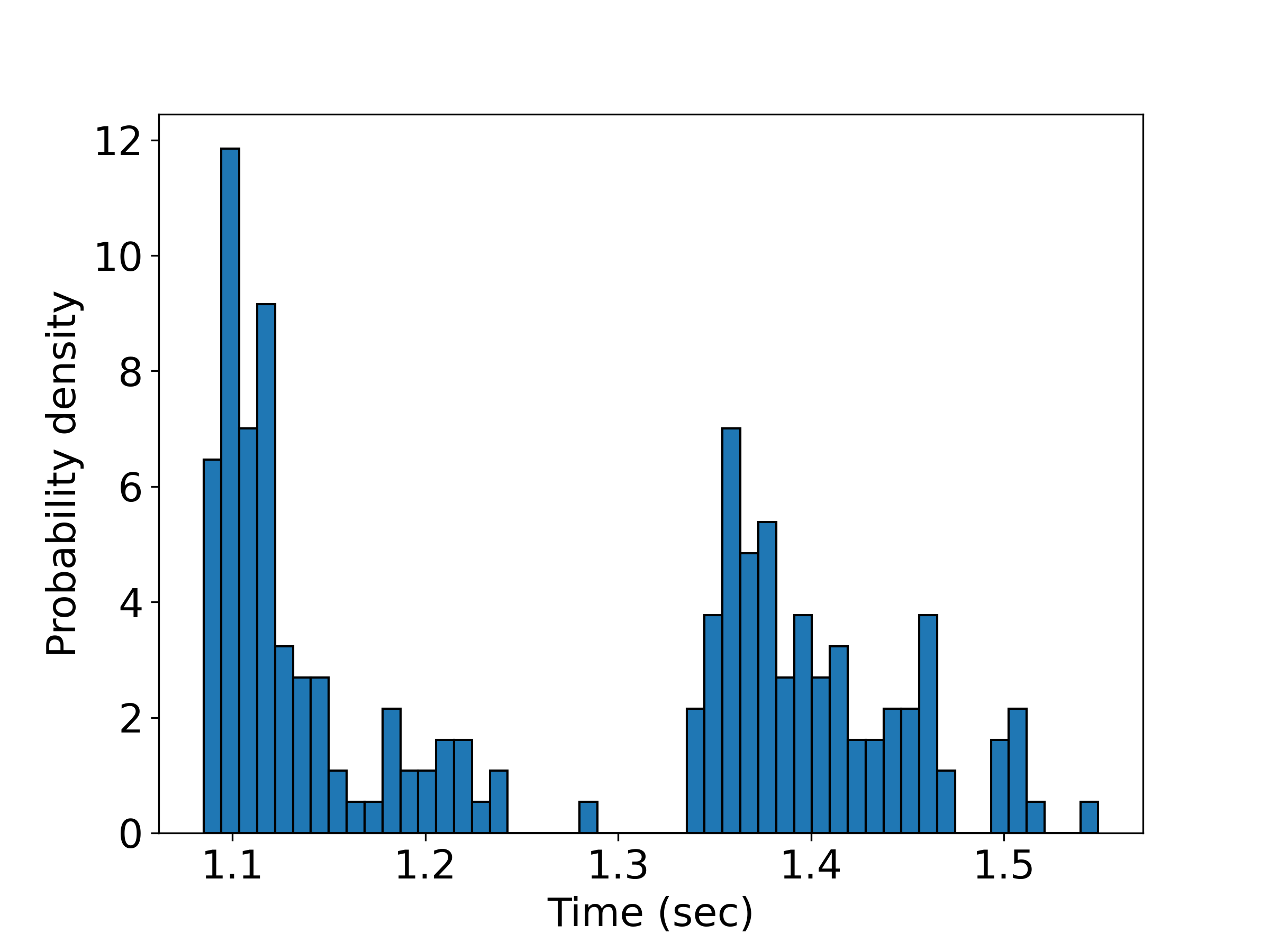}}
    \caption{top: GPU run time at different confidence threhsold, bottom: Distribution of  GPU run time of entity recognition model for CoNLL data set on 100 data set ( $\sigma=.15$ and $\mu=1.26$)}
    \label{fig:conll_time}
\end{figure}

GPU run time as figure \ref{fig:conll_time}a shows has greater fluctuation compared to the classification tasks e.g. figures \ref{fig:conll_time}b. 

The higher fluctuation in  GPU run time for this data set is related to the high variance of mamabear run time. As an example run time of 100 documents running 200 times shows  a noticeable variance.

\begin{figure}[h]
    \centering
    \subfloat{\includegraphics[width=1\linewidth]{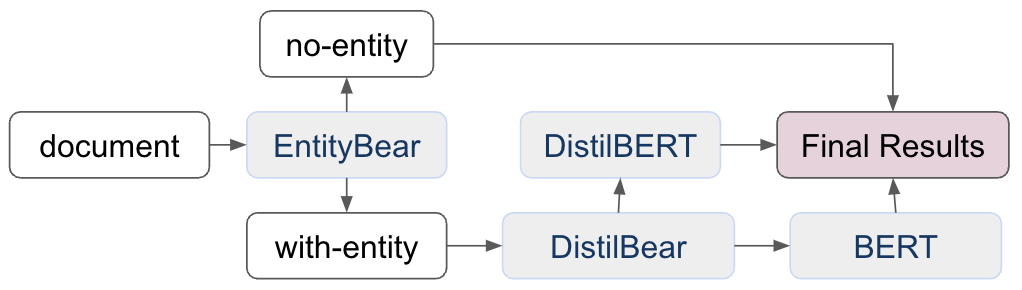}}\\
    \subfloat{\includegraphics[width=1\linewidth]{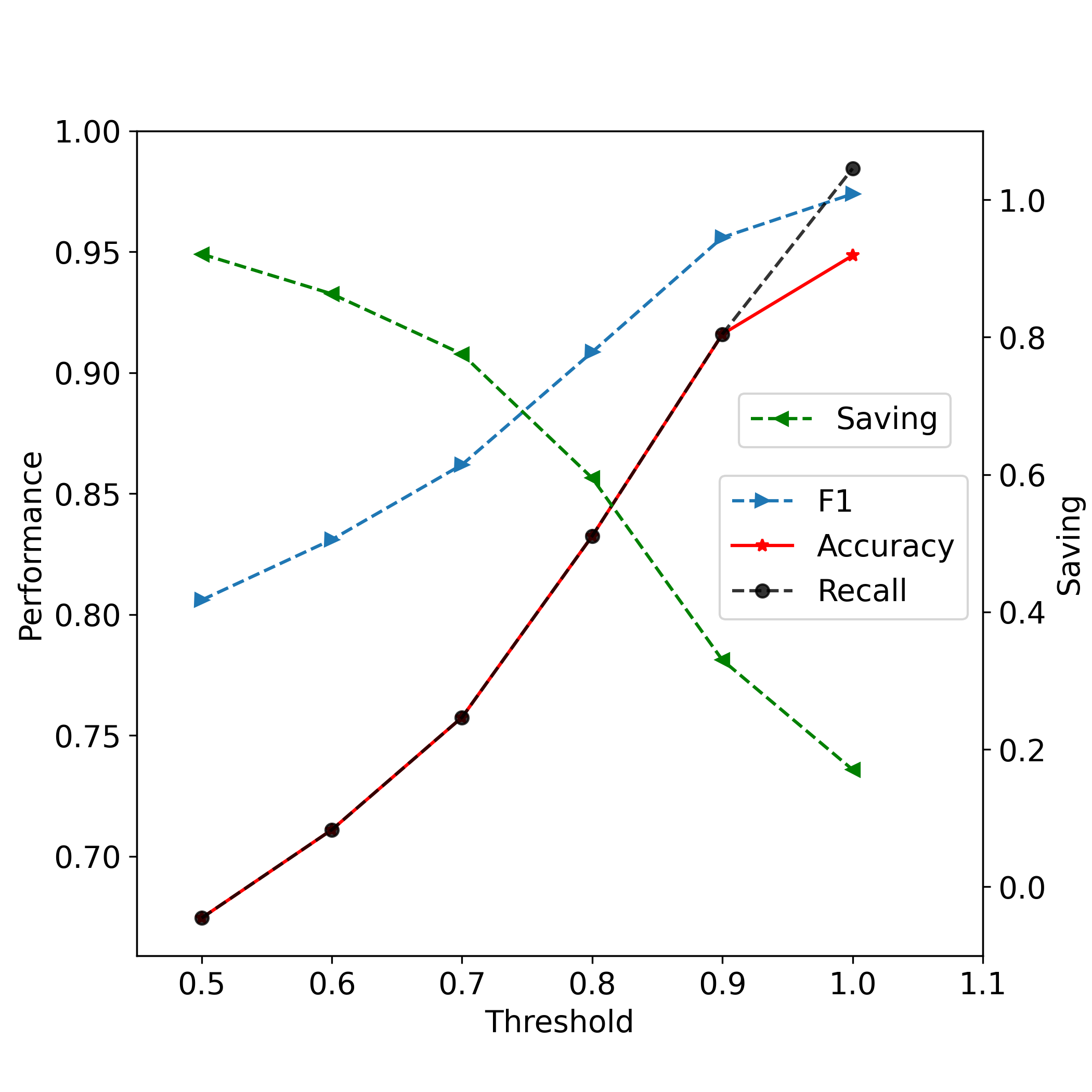}}
    \caption{top: DistilBear architecture. Input documents are first
    processed by EntityBear model. If  document is identified as "with-entity" it will predicted by DistilBERT model. Otherwise, the more accurate final BERT model processes the document and produces the final prediction. Bottom: Performance and saving of DistilBear model on CoNLL data set.}
    \label{fig:distilbear}
\end{figure}

\textbf{DistilBear.}
In the previous section we applied a simple babybear model on different data sets. For  complicated problems like entity recognition, this algorithm can be expanded into a cascade of multiple babybear models. 
Maximum saving using the proposed entity recognition model (EntityBear) is limited to the maximum number of  "no-entity" sentences. Here we proposed a combination of multiple babybear models. This algorithm is shown in figure \ref{fig:distilbear}.

EntityBear uses the model trained in section \ref{sec:ner}.  DistilBear uses a binary classification model trained on labels 0 and 1. Label  "1" is assigned when  DistilBERT model can correctly predict the entities in a sentence. Sentences predicted as class 0 will be sent to the BERT model. DistilBear is trained on 2000 data examples using an XGBoost classifier. This algorithm is applied on 2000 data examples. EntityBear saves 341 documents, while the rest are sent to DistilBear, the mamabear. Figure \ref{fig:distilbear} shows the performance of DistilBear for different threshold values. For high values of threshold, saving in limited to the savings on EntityBear. However for $t=.9$ by holding F1 score over than 95\% we can get to savings of 33\%  which is  higher than the potential saving in EntityBear model.

\section{Conclusion}

In this paper, we propose an inference triage framework, BabyBear, with the goal of accelerating the inference time for NLP tasks and reducing the cost of inference. Inference triage achieves those goals by using much less expensive models that run on CPU for the subset of tasks that they can confidently handle. Conventional (non-deep) machine learning models are shown to be adequate for a large proportion of inference examples for both classification and entity recognition tasks. A deep learning language model is only applied when the triage algorithm invokes it.

Our experimental results on two classification data sets (emotion recognition, sentiment analysis) and two entity recognition data sets (CoNLL, ncbi\_disease)  demonstrate substantial savings on inference time without significantly sacrificing performance. This algorithm reduced GPU run time for classification tasks by up to 64\%  while maintaining accuracy over 90\%.

We also explored variations on the BabyBear framework, expanding it into a cascade of multiple babybear models and applying it to entity recognition tasks. For the CoNLL data set, we maintained a 95\% of F1 score while reducing the GPU run time by 33\%.

We should stop using transformer language models as the only tool in our NLP toolkit. Let BabyBear help out.

\section{Appendices}

\section*{Acknowledgements}
We thank Oleg Vasilyev and Jake Asquith for review of the paper and valuable feedback.

\bibliography{anthology,custom}
\bibliographystyle{acl_natbib}

\end{document}